\documentclass{article}

    \PassOptionsToPackage{numbers, compress}{natbib}

\usepackage{times}
\usepackage{epsfig}
\usepackage{graphicx}
\usepackage{amsmath}
\usepackage{array}
\usepackage{amssymb}
\usepackage{comment}
\usepackage{xcolor}
\usepackage[font=scriptsize]{subfig}
\usepackage[font=small]{caption}
\usepackage{siunitx, mhchem}
\usepackage{booktabs}
\definecolor{Road}{RGB}{128, 64, 128}
\definecolor{Sidewalk}{RGB}{244, 35, 232}
\definecolor{Building}{RGB}{70, 70, 70}
\definecolor{Wall}{RGB}{102, 102, 156}
\definecolor{Fence}{RGB}{190, 153, 153}
\definecolor{Pole}{RGB}{153, 153, 153}
\definecolor{T. Light}{RGB}{250, 170, 30}
\definecolor{T. Sign}{RGB}{220, 220, 0}
\definecolor{Vegetation}{RGB}{107, 142, 35}
\definecolor{Terrain}{RGB}{152, 251, 152}
\definecolor{Sky}{RGB}{0, 130, 180}
\definecolor{Person}{RGB}{255, 127, 80}
\definecolor{Rider}{RGB}{255, 0, 0}
\definecolor{Car}{RGB}{0, 0, 142}
\definecolor{Truck}{RGB}{0, 0, 70}
\definecolor{Bus}{RGB}{0,206,209}
\definecolor{Train}{RGB}{0, 80, 100}
\definecolor{Motorcycle}{RGB}{100, 149, 237}
\definecolor{Bicycle}{RGB}{119, 11, 32}

\definecolor{Shuaib}{RGB}{250,125,0}

\usepackage[final]{neurips_2022}

\usepackage[utf8]{inputenc} 
\usepackage[T1]{fontenc}    
\usepackage{hyperref}       
\usepackage{url}            
\usepackage{booktabs}       
\usepackage{amsfonts}       
\usepackage{nicefrac}       
\usepackage{microtype}      
\usepackage{xcolor}         

\title{CLUDA : Contrastive Learning in Unsupervised Domain Adaptation for Semantic Segmentation}

\author{
    Midhun Vayyat, Jaswin Kasi, Anuraag Bhattacharya, Shuaib Ahmed, Rahul Tallamraju\\
    Mercedes Benz Research and Development India \\
\{midhun.vayyat, kasi.jaswin, anuraag.bhattacharya, shuaib.ahmed, rahul.tallamraju\}@mercedes-benz.com \\
}

\begin{document}

\maketitle

\begin{abstract}
  In this work, we propose CLUDA, a simple, yet novel method for performing unsupervised domain adaptation (UDA) for semantic segmentation by incorporating contrastive losses into a student-teacher learning paradigm, that makes use of pseudo-labels generated from the target domain by the teacher network. More specifically, we extract a multi-level fused-feature map from the encoder, and apply contrastive loss across different classes and different domains, via source-target mixing of images. We consistently improve performance on various feature encoder architectures and for different domain adaptation datasets in semantic segmentation. Furthermore, we introduce a learned-weighted contrastive loss to improve upon on a state-of-the-art multi-resolution training approach in UDA. We produce state-of-the-art results on GTA $\rightarrow$ Cityscapes (74.4 mIOU, +0.6, standard deviation: 0.32) and Synthia $\rightarrow$ Cityscapes (66.8 mIOU, +1.0, standard deviation: 0.44) datasets. CLUDA effectively demonstrates contrastive learning in UDA as a generic method, which can be easily integrated into any existing UDA for semantic segmentation tasks. Please refer to the appendix (section \ref{sec:code_impl}) for the details on implementation.
\end{abstract}
\section{Introduction}
Semantic segmentation is an important task in computer vision with applications ranging from autonomous driving to medical image analysis.  Most of the existing approaches in semantic segmentation \cite{liu2021domain, musto2020semantically, zhang2021prototypical, zhao2021contrastive, wang2021cross} use deep learning models in a fully supervised setting. This requires a lot of annotated data for accurate predictions. However, manual annotation of class labels is very difficult, as it may take around 1.5 hours of human labour to annotate a single driving scene image from the cityscapes dataset \cite{sakaridis2021acdc, hoyer2021daformer}.
One promising way to circumvent the issue of manual annotation is instead to train models on synthetically rendered, and auto-annotated images. High-resolution photo-realistic images can be generated from advanced computer graphics software/game engines \cite{richter2016playing, varol2017learning}. A major advantage is that a large amount of annotated data can be rendered in a relatively short amount of time. However, it has been demonstrated systematically in previous work that models trained only on synthetic data do not generalize well on real-world data \cite{zou2018unsupervised}. Despite being photo-realistic, synthetically generated data suffers from domain shift, as the underlying data distribution of real and synthetic images are different. Hence, using models trained on synthetic data (source domain) for real-world applications would require sophisticated domain adaptation strategies to generalize to the real-world (real domain).

Extensive research on domain adaptation has been carried out in the literature. In scenarios where manual annotations are available, along with synthetic data, semi-supervised and weakly-supervised domain adaptation methods have been explored \cite{french2019semi, hoyer2021three, lai2021semi, souly2017semi, dai2015boxsup, song2019box, zou2020pseudoseg}. However, when the manual annotation is difficult or tedious to obtain, unsupervised domain adaptation (UDA) techniques have been proposed \cite{hoffman2016fcns, tsai2018learning, zou2018unsupervised}. 

Generally, UDA is performed in two steps. Models are first trained on source domain data (synthetic data). These models are subsequently used to generate the pseudo-labels on the target domain data (real-world data). Since the pseudo-labels are noisy and the learnt features for the two domains are distinct, a second step in the form of adversarial training  \cite{hoffman2016fcns, tsai2018learning} or self-learning \cite{zhang2021prototypical, zou2018unsupervised} is employed. The general architecture of such models is as follows. An encoder with feature extractor backbone (ResNet-101\cite{he2016deep}, VGG-19\cite{simonyan2014very}, and more recently SegFormer\cite{hoyer2021daformer}), followed by a classifier that predicts dense labels. Most recent domain adaptation methods for semantic segmentation like DAFormer \cite{hoyer2021daformer}, HRDA \cite{hoyer2022hrda} use transformer backbones to extract features and a cross-entropy loss to train the classifier. Intuitively cross-entropy loss aims only at bringing similar features together while ignoring to differentiate features across distinct classes. Hence, it is important to separate features corresponding to different classes while accumulating features belonging to the same class in the latent space.

In this paper, we introduce contrastive learning along with cross-entropy loss for unsupervised domain adaptation called \textit(CLUDA) to keep similar features in latent space together while separating away distinct features. This results in a richer, compact and well separated feature space, thereby, making it easier for final classification. Based on the above intuition we argue that intra-class compactness and inter-class separability of features in source domain remain similar in the target domain. 
Contrastive learning has been recently applied for semantic segmentataion in both supervised \cite{hu2021region, zhong2021pixel, zhang2021looking, xie2021propagate, wang2021dense, lai2021semi, wang2021cross} and unsupervised settings \cite{he2020momentum,chen2020simple,cai2020joint,hadsell2006dimensionality,oord2018representation}.In this work, we demonstrate that the proposed method can be used to enhance the domain adaptation capabilities for semantic segmentation on state-of-the-art encoder architectures. We showcase state-of-the-art results by applying contrastive losses to transformer-based UDA methods, like DAFormer and HRDA.

\textbf(Contributions:) We propose the usage of contrastive losses for unsupervised domain adaptation in semantic segmentation. Our method increases intra-class semantic similarity and decreases inter-class similarity across source and target domains. We also show that our proposed approach outperforms the current state-of-the-art on different datasets such as GTA $\rightarrow$ Cityscapes and Synthia $\rightarrow$ Cityscapes, where $\rightarrow$ indicates mapping from source to target domain datasets. Moreover, to overcome the shape-bias problem in transformers, we propose a novel masking strategy for stuff and thing classes separately. We also demonstrate the flexibility of our method by applying it to two state-of-the-art UDA approaches DAFormer \cite{hoyer2021daformer} and multi-resolution HRDA \cite{hoyer2022hrda}.

\section{Related works}
\subsection{Semantic Segmentation}
Most of the existing works in semantic segmentation use per-pixel cross-entropy loss between the predicted label and ground truth, as first shown in \cite{long2015fully}. Besides cross-entropy loss and its variants \cite{xie2015holistically, abraham2019novel}, Dice loss has been used, mainly to detect the boundaries of the objects in the image for better segmentation \cite{diakogiannis2020resunet, zhang2017brain, soomro2018strided}. Also, there are few works on feature-distribution-based losses, as introduced in \cite{wang2021cross, zhao2021contrastive}.

\subsection{UDA in Semantic Segmentation} 
Some methods \cite{hoffman2018cycada, musto2020semantically} explores domain adaptive
segmentation along with image-to-image translation using cyclic
loss \cite{zhu2017unpaired}. Since the target annotation is not available, to generate pseudo
label, a hard threshold is used to eliminate low confidence pixel predictions
from the predicted label. The work \cite{liu2021source} proposes source-free domain adaptation using a model pre-trained only on source dataset. This model will serve as a source dataset generator, which later helps in domain adaptation to target data, via adversarial loss training. 

Recently \cite{liu2021domain} proposed a patch-wise intra-image contrastive learning for both semi-supervised and unsupervised domain adaptation. Images are divided into patches, and features extracted from these patches are used in contrastive learning. In \cite{wang2021domain}, depth estimation is used as an auxiliary task to alleviate the domain shift. It uses the depth estimation discrepancies in two domains to refine the target pseudo-labels and guide the domain adaptation. In another recent work \cite{zhang2021prototypical}, the pseudo-label predictions were rectified exploiting the distance between features and their respective class centroids. Furthermore \cite{wang2021cross} improves upon \cite{zhang2021prototypical} by aligning untrusted and trusted pixel regions using adversarial training. Along with cross-domain adaptation, one can perform cross-region adaptation in this manner. Broadly speaking, to improve the feature extraction capability of the encoder we explore the contrastive learning approach which has shown some promising results recently.

\subsection{Contrastive Losses for Semantic Segmentation}
 In \cite{chopra2005learning}, Sumit et. al. invented the concept of
contrastive learning by bringing similar samples together, while separating different labels apart. Florian et. al. \cite{schroff2015facenet} introduced triplet-loss which takes a sample and brings it closer to its class anchor called positive anchor and at the same time drives it farther from other class anchors called negative anchors. Contrastive loss techniques can also be differentiated based on the kind of metric-learning that is used to cluster the features.
Some works \cite{chopra2005learning, schroff2015facenet} use Euclidean distance to compute the distance between two features, while others
\cite{zhao2021contrastive, wang2021cross} use cosine similarity between the unit-normalized features to align similar features and separate
dissimilar features apart. There are quite a few variants of contrastive losses(CL), like region-aware CL\cite{hu2021region}, positive-negative equal CL\cite{wang2022positive}, intra-image pixel-wise CL\cite{zhong2021pixel}, cross-image pixel-wise CL\cite{wang2021cross}, centroid-aware CL, and distribution-aware CL\cite{xie2022sepico}.

Contrastive learning in semantic segmentation have only been recently introduced in \cite{hu2021region, zhong2021pixel, zhang2021looking, xie2021propagate, wang2021dense, lai2021semi, wang2021cross}. Since semantic segmentation is a dense prediction task each pixel is a feature and the cluster representation is learned at pixel level. While \cite{wang2021exploring} exploits the dense contrastive learning in a fully supervised setting, \cite{lai2021semi} takes a semi-supervised approach and \cite{xie2021propagate, wang2021dense} follows the unsupervised approach. Similar to \cite{wang2021exploring} we exploit the inter-image semantic similarity and dissimilarity between classes but extend it to unsupervised domain adaptation. These works attempt to align the same class features and distinguish different class features in the feature space. All of these works use the feature generated at the last layer of the feature extractor. We attempt to combine the features using contrastive losses by using all the features generated by the hierarchical features extractors like \cite{he2016deep, xie2021segformer, liu2021swin}. 
Recent works have focused on unsupervised contrastive learning \cite{he2020momentum,chen2020simple,cai2020joint,hadsell2006dimensionality,oord2018representation} because of its capability to learn feature representation in the absence of human supervision. We define unsupervised contrastive loss objective in section \ref{sec:da}. We utilize the existing InfoNCE\cite{gutmann2010noise} loss for supervised contrastive loss for source training. For domain adaptation we modify the same loss by hinging it on model prediction confidence.

\section{Prerequisites}\label{sec:prereqs}
Given a source (synthetic) dataset $\{X_s = x_{i=1}^{n}\}$ with segmentation labels $\{Y_s = y_{i=1}^{n}\}$, we train a deep learning model that learns from the source and is expected to produce high accuracy semantic maps for the target (real) images $\{X_t = x_{i=1}^{m}\}$ without annotation data. Here, both $X_s$ and $X_t$ share same $C$ classes. Generally a semantic segmentation model follows an encoder-decoder architecture, where encoder is a feature extractor $f$ followed by a decoder that has a classifier $g$. A semantic segmentation map is given by $h = g(f(x))$, which predicts classes for each pixel in the image. The network is optimised using categorical cross-entropy $l_{ce}$ with ground-truth and pseudo ground-truth for real and synthetic data respectively. 
\begin{equation}
\footnotesize
l_{ce}^{s} = - \sum \limits_{i=1}^{H \times W} \sum \limits_{c=1}^{C} y_s^{(i,c)} \log(p_s^{(i,c)}), \qquad l_{ce}^{t} = - \sum \limits_{i=1}^{H \times W} \sum \limits_{c=1}^{C} \hat{y}^{(i,c)}_t \log(p_t^{(i,c)})
\end{equation}
where $y_s^{(i,c)}$ is the source ground-truth label, $\hat{y}_t^{(i,c)}$ is the target pseudo- label and $p_t^{(i,c)}$ is the softmax probability of $x_t^{(i)}$ pixel belonging
to class $c \in C $. 
Since we cannot use real ground- truth we use self-training \cite{zhang2021prototypical, zhu2020improving, hoyer2021daformer, wang2021domain} to infer pseudo-labels. 

For self-training with only cross-entropy loss, we adopt the approach taken in DAFormer \cite{hoyer2021daformer}. It consists of a student-teacher network, where pseudo-labels of target images are predicted by the teacher network $h_t$ with the following formulation - $p_t^{(i,c)} = \left[i = argmax(h_t(x_t)_c)\right]$, where $\left[. \right]$ is the Iversion bracket, 
Whereas student network sees a source-target-mixed-image.
A mixed-ground-truth for the mixed-image is then constructed by overlaying the source ground-truth(properly aligning as per the mixing) on pseudo-labels predicted by teacher network. Using this mixed-ground-truth the student network is trained, while freezing the teacher network. Teacher network is indirectly updated via exponential-moving-average (EMA) of the weights from the student network. Following DAFormer, we control EMA update weights using a fixed hyper-parameter $\alpha$, and as well use, rare class sampling (RCS) to alleviate the issue of long-tail representation of the classes in the source data. Moreover, we use ImageNet \cite{deng2009imagenet} feature distance loss to prevent the network from over-fitting to the source data. For finer details we redirect the readers to the aforementioned paper.
\begin{figure*}
\begin{center}
\includegraphics[width=0.9\linewidth]{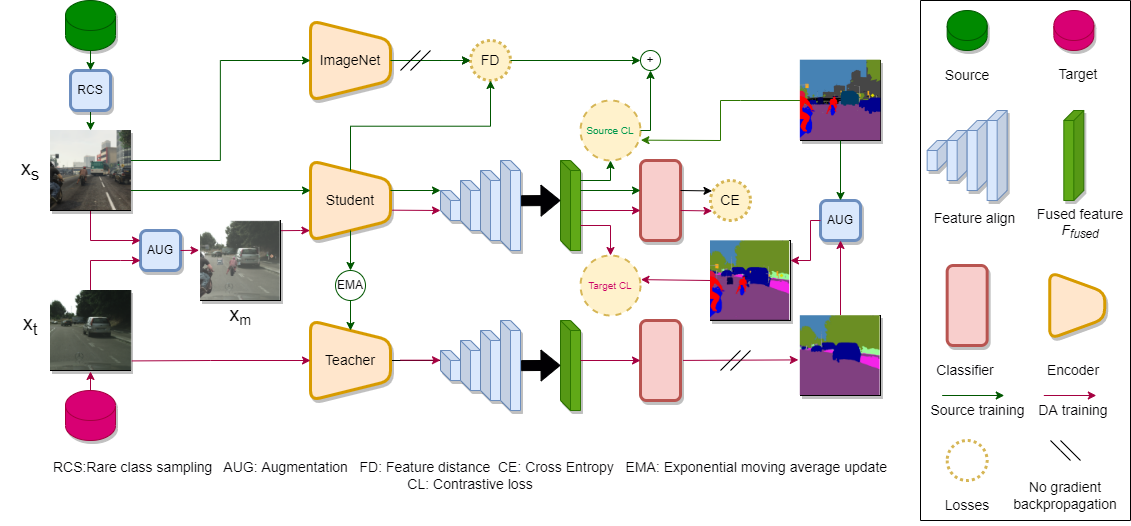}           
\end{center}
\scriptsize
\caption{We sample source images using rare class sampling (RCS) and use a fixed ImageNet pre-trained encoder for preventing the student network from forgetting real features following \cite{hoyer2021daformer}. We fuse the encoder hierarchical features and compute contrastive losses on the fused features. For DA training we use the self learning approach to generate the pseudo-label but hinge the contrastive loss computed for domain mixed images on the teacher net prediction confidence.}
\label{fig:full_model}
\end{figure*}

\section{Our Approach}
Building upon motivations described in previous sections, we introduce our approach which uses contrastive loss for UDA in semantic segmentation. We start by providing an overview of our proposed approach. In the subsequent subsections, we describe contrastive loss used (section \ref{sec:cl}), training with source domain data (section \ref{sec:Contrastive Source Training}), and finally domain adaptation to target domain data (section \ref{sec:da}). Additionally we introduce contrastive loss in Multi-Resolution DA setting (section \ref{sec:multi_res_da}).

\subsection{Overview of our approach}
 Figure \ref{fig:full_model} provides the schematic of our proposed approach. The input to our model is a source image and a target image sampled from the source and target distributions respectively. We subsequently create a mixed image, where the source and target images are mixed using the approach in ClassMix\cite{olsson2021classmix}. Here source pixels belonging to randomly sampled classes are overlayed on the target image and pixel level augmentations are applied to the resulting mixed image. The source image $x_s$ is first passed through the student model, which generates a multi-resolution feature map. These feature maps are fused by a CNN based fusion network. This fused feature map is passed through a classifier to generate pixel level predictions. We apply contrastive loss (Source CL) on the fused feature map and cross entropy (CE) loss on the classifier output. We additionally pass the image through an ImageNet \cite{deng2009imagenet} pretrained model and compute the feature-distance (FD) loss between the student model and the ImageNet pretrained model features as in \cite{hoyer2021daformer}. Details of the contrastive loss computation on source features is provided in section \ref{sec:cl}. Similarly the target image $x_t$ is passed through the teacher model. The teacher model follows the same architecture as the student model, however its weights are updated as an EMA of the student weights. The prediction of the teacher model acts as a pseudo-label for the target image. Finally, the mixed image $x_m$ is also passed through the student model. These predictions are supervised using the original labels for source pixels and the pseudo-labels for the target pixels of the mixed image. We apply contrastive loss (Target CL) on the fused feature map and cross entropy (CE) loss on the classifier output. Details of the contrastive loss computation on mixed features is provided in section \ref{sec:da}.

\subsection{Contrastive Losses} \label{sec:cl}
 In this paper, we follow the contrastive loss defined in \cite{wang2021exploring}. For a given pixel $i$, the corresponding feature-vector is represented by $\vec{v}_i$, the contrastive loss $\mathcal{L}^{I}_{NCE}$ is then given by, \ref{InfoNCE},
\begin{equation}
\label{InfoNCE}
\footnotesize
\mathcal{L}^{I}_{NCE}(\vec{v}, \mathcal{P}, \mathcal{N}) = \frac{1}{\left| \mathcal{P}_i \right|} \sum\limits_{v^{+} \in \mathcal{P}_i} \Biggl[ -\log \biggl( \dfrac{e^{(\vec{v}_i.\vec{v}^{+}/\tau)}}{ e^{(\vec{v}_i.\vec{v}^+ / \tau)} + \sum\limits_{v^- \in \mathcal{N}_{i}} e^{(\vec{v}_i.\vec{v}^{-}/\tau)}} \biggr) \Biggr].
\end{equation}
Here, $\vec{v}^+$ represents a positive anchor (feature vector from the same class), $\vec{v}^-$ is the negative anchor (feature vector from a different class) and $\tau$ is the temperature hyper-parameter to control the magnitude of the similarity. The summation is normalized by the cardinality of all the positive samples considered for the pixel $i$, indicated by $|\mathcal{P}_i|$. Similarly, $\mathcal{N}_i$ is the set of all negative samples considered. 
Following \cite{wang2021exploring, zhao2021contrastive}, we compute the loss between every pair of pixel-features $\vec{v}_{i}, \vec{v}^{+/-}$ across the batch.

\subsection{Contrastive Source Training}\label{sec:Contrastive Source Training}
Feature extractors like ResNet-101 \cite{he2016deep}, SegFormer \cite{xie2021segformer} produce hierarchical feature maps at different scales with different embedding dimensions. Current works compute contrastive losses over the final feature map, and neglect the rest of the output feature maps (see Fig. \ref{fig:full_model}). However, the hierarchical feature maps capture high resolution fine features and low resolution coarse features \cite{xie2021segformer, he2016deep}. In our approach, we propose to make use of all the feature maps generated by the feature extractor for contrastive loss computation. In DAFormer \cite{hoyer2021daformer},
the generated feature maps are fused after channel and shape aligning, to produce a single fused-feature tensor $\mathcal{F}_{fused}$. $\mathcal{F}_{fused}$ is used to compute $N$-channel segmentation logits. $\mathcal{F}_{fused}$ tensor for a given image is of size $H/\mu \times W/\mu \times C$, where $H,W$ are height and width of the image, $\mu$ is a down-scaling factor that defines the output feature map size. $\mathcal{A} = (H/\mu \times W/\mu)$ is a normalizing factor. $C$ is the embedding size of pixel-feature vectors $\vec{v}_i$.  $\mathcal{P}_{s} / \mathcal{N}_{s}$ indicates positive/negative samples from both source and target pixels in $x_s$ respectively. Due to GPU memory constraints, following \cite{zhao2021contrastive}. We compute the contrastive loss for source images using equation \ref{InfoNCE}. The anchor pixels are chosen from the ground-truth labels of the source images.
\begin{equation}
\label{eq:src_CLUDA}
\mathcal{L}_{CL}^{s} = \dfrac{1}{\mathcal{A}} \biggl[ \sum\limits_{s \in \mathcal{S}}\mathcal{L}^{I}_{NCE}(\vec{v}_s, \mathcal{P}_{s}, \mathcal{N}_s) \biggr].
\end{equation}
\subsection{Contrastive Domain Adaptation Training}\label{sec:da}
\begin{figure}[h!]
\hspace{-1.0cm}\subfloat[\label{fig:multi_res_cluda}]{
    \includegraphics[width=0.6\linewidth]{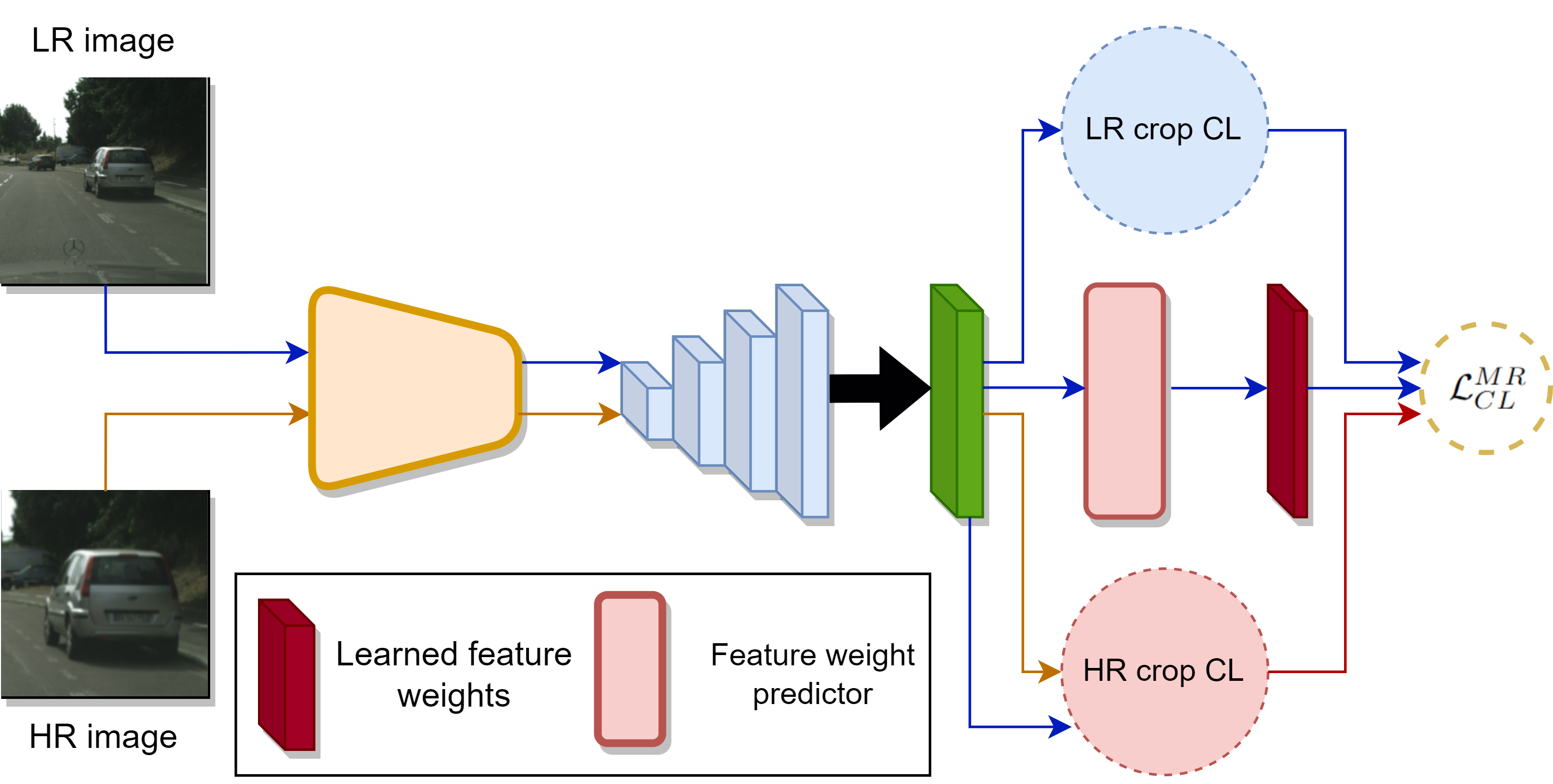}
}
\subfloat[\label{gph:pixel_rate}]{
    \includegraphics[width=0.25\linewidth]{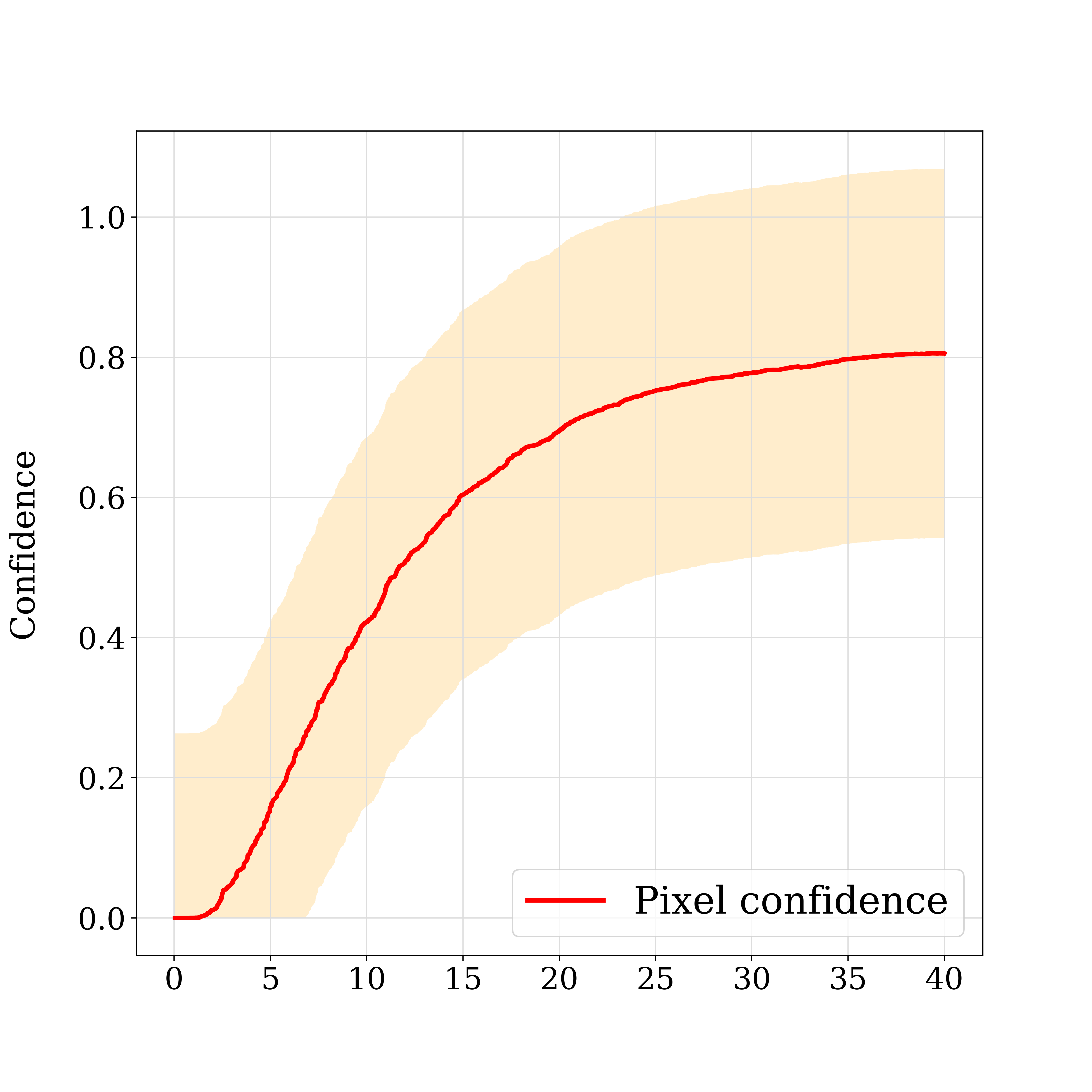}
}
\subfloat[\label{gph:weighting_perf_comp}]{
    \includegraphics[width=0.25\linewidth]{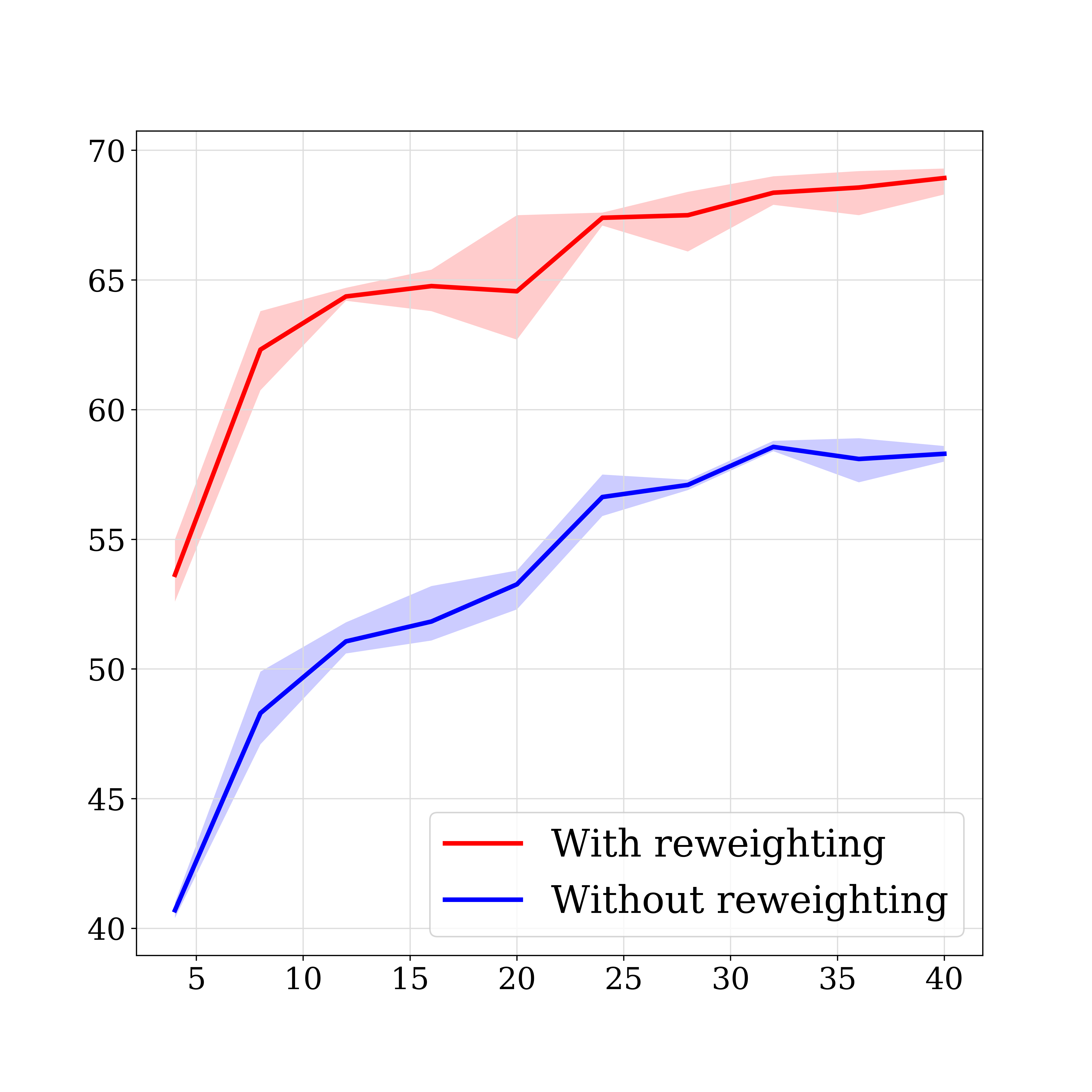}
}
\caption{Fig (a) shows the CLUDA module for multi-resolution UDA. Note that the figures shows just the contrastive learning module using multi-resolution, rest of the architecture remains same as HRDA\cite{hoyer2022hrda} and that we have mentioned only those elements in the legend that are new in this architecture, rest of the elements are same as Fig. \ref{fig:full_model}. Please refer to equation \ref{multi_res_cl} for the loss $\mathcal{L}_{CL}^{MR}$. Fig. (b) and (c) shows the pixel confidence and improvement in the mIoU score when the contrastive losses applied on the domain-mixed features are weighted according to the confidence score computed using equation respectively \ref{eq:conf_computation}.}
\label{confidence_improvement_graph}
\end{figure}

We follow \cite{tranheden2021dacs} to do DA training, where ClassMix \cite{olsson2021classmix} is used to create mixed images ($x_m$ in Fig. \ref{fig:full_model}). Additionally pixel level augmentations like color-jitter, blur and saturation are applied. The mixed image $x_m$ enables the network to view pixels from the source and target domain simultaneously, and thereby aids in domain adaptation. 
By applying contrastive losses on the features extracted from the mixed image, we aim to align features belonging
to the same class from both the domains, and separate out features belonging to different classes. In UDA, the target semantic labels are not available so we rely on the pseudo-labels (as previously described in section \ref{sec:prereqs}). In CLUDA, for the mixed images, the contrastive losses are weighted based on teacher network pseudo-label confidence on the target pixels. Without such re-weighting, the model produces inconsistent results as can be seen in Fig. \ref{gph:weighting_perf_comp} (see section \ref{sec:conf_weight_sec}). For computing the weighing-factor $\Gamma$, we first compute the number of target-pixels whose softmax-probabilities $P$ are above a certain threshold $\beta$. 
In equation \ref{eq:conf_computation}, $T$ denotes all the pixels in the target image, $i$ denotes the pixel, $k$ denotes the class. Then the contrastive loss is given by,

\begin{equation}\label{eq:conf_computation}
\Gamma = \dfrac{1}{\mathcal{A}} \sum_{i \in T} 1_{[max_k P_{i,k} > \beta]}, 
\end{equation}
\begin{equation}
\label{eq:CLUDA}
\mathcal{L}_{CL}^{m} = \dfrac{1}{\mathcal{A}} \biggl[ \sum\limits_{s \in \mathcal{S}}\mathcal{L}^{I}_{NCE}(\vec{v}_s, \mathcal{P}_{s}, \mathcal{N}_s) + \Gamma \sum\limits_{t \in \mathcal{T}} \mathcal{L}^{I}_{NCE}(\vec{v}_t, \mathcal{P}_{s, t}, \mathcal{N}_{s, t}) \biggr].
\end{equation}

Here, the overall loss consists of two terms. The first term is for source feature vectors $\mathcal{S}$ in $x_m$ whereas the second term is for target feature vectors $\mathcal{T}$ in $x_m$. Note that the contribution of target feature vectors has been down weighted by the confidence factor $\Gamma$. $\mathcal{P}_{s, t} / \mathcal{N}_{s,t}$ indicates positive/negative samples from both source and target pixels in $x_m$ image respectively, i.e while computing the contrastive loss, each term considers source and target feature vectors. We observe that the pseudo-label prediction confidence increases over the course of the training as shown in Fig. \ref{confidence_improvement_graph},
as the model gets exposed to more source and target pixels. The losses $\mathcal{L}_{CL}^{s}$ and $\mathcal{L}_{CL}^{m}$ is simply added to the losses mentioned in Section 3.3 of DAFormer \cite{hoyer2021daformer}.
\begin{equation}
\mathcal{L} =  l^s_{ce} + l^t_{ce} + \mathcal{L}_{CL}^{s} + \mathcal{L}_{CL}^{m} + \lambda \mathcal{L}_{FD}
\end{equation}
where $\mathcal{L}_{FD}$ is the Euclidean-loss between student-network features and fixed ImageNet-features on only source images. Henceforth we refer to this addition as \textbf{DAFormer \cite{hoyer2021daformer} + CLUDA}. \textbf{Note}, for discussions in subsection \ref{Influence of stuff-thing mask}, we compute contrastive loss of stuff-classes and thing-classes differently. For the definition of stuff- and thing-classes we follow \cite{caesar2018coco}. We don't include comparisons between stuff and thing-classes feature vectors in the equation \ref{eq:CLUDA}.
\subsection{Multi-Resolution CLUDA}\label{sec:multi_res_da}

Recently multi-resolution images in UDA was introduced by HRDA\cite{hoyer2022hrda} to improve performance of finer classes such as Traffic Light, Traffic Sign, etc. HRDA uses a low-resolution (LR) image that captures the long-range contextual dependencies, and a high-resolution (HR) image-crop that helps in predicting high-quality segmentation maps of the smaller-classes. The $N$-class prediction logits of both LR and HR are fused using a learned scale-attention. For UDA, the target pseudo-label is generated by fusing together segmentation prediction (from the teacher model \cite{hoyer2022hrda}) of overlapping sliding window crops of the target image. These pseudo-labels along with source ground-truth labels are used during domain adaptation.

Introducing contrastive loss naively in HRDA, turns out to be ineffective (see Table \ref{multi_res_comp}). We posit that the representations learnt for LR and HR crops are different, and hence training with CL over just the LR features is not sufficient, as the CE loss will force the model to learn from both LR and HR predictions. Therefore, we propose a novel strategy to utilize both the LR and HR features for CLUDA. We combine the per-pixel CL from LR and HR image crops, using a learned scale weight for each pixel. Note that the learned scale weights are computed on LR crop only. Effectively, these weights quantify the confidence for considering features from high-resolution crops to compare against features from low-resolution. Contrastive loss in equation \ref{eq:CLUDA} is replaced by, 
\begin{equation}
\label{multi_res_cl}
\begin{split}
& \mathcal{L}_{CL}^{MR}(x^{l}_m, x^{h}_m, \mathcal{A}^{s}, \mathcal{A}^{m}) = 
\frac{1}{\mathcal{A}} \biggl[ \sum\limits_{s \in \mathcal{S}_l} \delta_l \mathcal{L}^{I}_{NCE}(\vec{v}^{l}_s, \mathcal{P}^{l}_{s}, \mathcal{N}^{l}_s) + \sum_{s\in S_h} (1 - \delta_l)  \mathcal{L}_{NCE}^I (\vec{v}^{h}_s, \mathcal{P}^{h}_{s}, \mathcal{N}^{h}_{s} ) \\ 
& \qquad + \Gamma \sum_{t \in \mathcal{T}_l} \delta_{l} \mathcal{L}^{I}_{NCE}(\vec{v}^{l}_{t}, \mathcal{P}^{l}_{s,t}, \mathcal{N}^{l}_{s,t}) + \Gamma \sum_{t\in \mathcal{T}_h} (1 - \delta_l)  \mathcal{L}_{NCE}^I (\vec{v}^{h}_{t}, \mathcal{P}^{h}_{s,t}, \mathcal{N}^{h}_{s,t} ) \biggr].  
\end{split}
\end{equation}
Here, $\delta_l$ is the learnable weightage per pixel computed on LR crop. $\mathcal{S}_{l,h}$ and ${\mathcal{T}_{l,h}}$ corresponds to LR and HR features of source and target pixels respectively. $\mathcal{N}^{h}/\mathcal{P}^{h}$ represent negative/positive features respectively from HR crop, while $\mathcal{N}^{l}/\mathcal{P}^{l}$ represent LR crop's negative/positive features respectively.  Note that, the weightage is $1$ for pixels present in LR crop and not available in the HR crop. Also note, we don't compare HR features against each other, i.e there is no CL terms of two HR feature-vectors in the loss function, in equation \ref{multi_res_cl}.

\section{Experiments}





\subsection{Implementation details}
\textbf{Datasets}: We perform our experiments on \textbf{GTA-5}\cite{richter2016playing}, \textbf{SYNTHIA}\cite{ros2016synthia}, \textbf{Cityscapes}\cite{cordts2016Cityscapes} datasets. Please refer to appendix for details on image resolution and dataset size.

\textbf{Network Architecture}: For DAFormer\cite{hoyer2021daformer} + CLUDA, we use Swin-L\cite{liu2021swin} as the backbone, while for HRDA\cite{hoyer2022hrda} + CLUDA, we use Segformer-B5\cite{xie2021segformer}. For both sets of experiments, we use a simple MLP decoder\cite{hoyer2021daformer} with embedding dimension of 512. The same decoder is used in the segmentation head, scaled attention head as well as feature weight prediction head in HRDA\cite{hoyer2022hrda} + CLUDA.
\begin{table}[h!]
\scriptsize
\setlength\tabcolsep{2pt}
\centering
\hspace*{-1.5cm}\begin{tabular}{ccccccccccccccccccccc}
    \toprule
    & Road & S.Walk & Build. & Wall & Fence & Pole & T. Light & T. Sign & Veget. & Terrain & Sky & Person & Rider & Car & Truck & Bus & Train & M.Bike & Bike & mIoU \\ \midrule
    \multicolumn{21}{c}{GTA $\rightarrow$ Cityscapes} \\
    \midrule
    AdaptSeg\cite{tsai2018learning} & 86.5 & 25.9 & 79.8 & 22.1 & 20.0 & 23.6 & 33.1 & 21.8 & 81.8 & 25.9 & 75.9 & 57.3 & 26.2 & 76.3 & 29.8 & 32.1 & 7.2 & 29.5 & 32.5 & 41.4\\
    CBST\cite{zou2018unsupervised} & 91.8 & 53.5 & 80.5 & 32.7 & 21.0 & 34.0 & 28.9 & 20.4 & 83.9 & 34.2 & 80.9 & 53.1 & 24.0 & 82.7 & 30.3 & 35.9 & 16.0 & 25.9 & 42.8 & 45.9 \\
    DACS\cite{tranheden2021dacs} & 89.9 & 39.7 & 87.9 & 30.7 & 39.5 & 38.5 & 46.4 & 52.8 & 88.0 & 44.0 & 88.8 & 67.2 & 35.8 & 84.5 & 45.7 & 50.2 & 0.0 & 27.3 & 34.0 & 52.1 \\
    CorDA\cite{wang2021domain} & 94.7 & 63.1 & 87.6 & 30.7 & 40.6 & 40.2 & 47.8 & 51.6 & 87.6 & 47.0 & 89.7 & 66.7 & 35.9 & 90.2 & 48.9 & 57.5 & 0.0 & 39.8 & 56.0 & 56.6\\
    BAPA\cite{liu2021bapa} & 94.4 & 61.0 & 88.0 & 26.8 & 39.9 & 38.3 & 46.1 & 55.3 & 87.8 & 46.1 & 89.4 & 68.8 & 40.0 & 90.2 & 60.4 & 59.0 & 0.0 & 45.1 & 54.2 & 57.4\\
    ProDA\cite{zhang2021prototypical} & 87.8 & 56.0 & 79.7 & 46.3 & 44.8 & 45.6 & 53.5 & 53.5 & 88.6 & 45.2 & 82.1 & 70.7 & 39.2 & 88.8 & 45.5 & 59.4 & 1.0 & 48.9 & 56.4 & 57.5 \\
    DAFormer\cite{hoyer2021daformer} & 95.7 & 70.2 & 89.4 & 53.5 & 48.1 & 49.6 & 55.8 & 59.4 & 89.9 & 47.9 & 92.5 & 72.2 & 44.7 & 92.3 & 74.5 & 78.2 & 65.1 & 55.9 & 61.8 & 68.3 \\
    DAFormer + \textbf{CLUDA} & \textbf{97.5} & \textbf{78.8} & 88.8 & 60.8 & 52 & 47.1 & 51.9 & 50.3 & 89.7 & 51 & 94 & 71 & 48.6 & 93.1 & 82 & 84.1 & 71.4 & 58.9 & 60.7 & \textbf{70.11} \\
    HRDA \cite{hoyer2022hrda} & 96.4 & 74.4 & \textbf{91} & \textbf{61.6} & 51.5 & \textbf{57.1} & 63.9 & 69.3 & \textbf{91.3} & 48.4 & 94.2 & \textbf{79.0} & \textbf{52.9} & 93.9 & \textbf{84.1} & 85.7 & \textbf{75.9} & 63.9 & 67.5 & 73.8 \\
    HRDA + \textbf{CLUDA} & 97.1 & 78 & \textbf{91} & 60.3 & \textbf{55.3} & 56.3 & \textbf{64.3} & \textbf{71.5} & 91.2 & \textbf{51.1} & \textbf{94.7} & 78.4 & \textbf{52.9} & \textbf{94.5} & 82.8 & \textbf{86.5} & 73 & \textbf{64.2} & \textbf{69.7} & \textbf{74.4}  \\
    \midrule
    \multicolumn{21}{c}{Synthia $\rightarrow$ Cityscapes} \\
    \midrule
    AdaptSeg\cite{tsai2018learning} & 79.2 & 37.2 & 78.8 & - & - & - & 9.9 & 10.5 & 78.2 & - & 80.5 & 53.5 & 19.6 & 67.0 & - & 29.5 & - & 21.6 & 31.3 & 37.2 \\
CBST\cite{zou2018unsupervised} & 68.0 & 29.9 & 76.3 & 10.8 & 1.4 & 33.9 & 22.8 & 29.5 & 77.6 & - & 78.3 & 60.6 & 28.3 & 81.6 & - & 23.5 & - & 18.8 & 39.8 & 42.6 \\
DACS\cite{tranheden2021dacs} & 80.6 & 25.1 & 81.9 & 21.5 & 2.9 & 37.2 & 22.7 & 24.0 & 83.7 & - & 90.8 & 67.5 & 38.3 & 82.9 & - & 38.9 & - & 28.5 & 47.6 & 48.3 \\
CorDA\cite{wang2021domain} & 93.3 & 61.6 & 85.3 & 19.6 & 5.1 & 37.8 & 36.6 & 42.8 & 84.9 & - & 90.4 & 69.7 & 41.8 & 85.6 & - & 38.4 & - & 32.6 & 53.9 & 55.0 \\ 
BAPA\cite{liu2021bapa} & 91.7 & 53.8 & 83.9 & 22.4 & 0.8 & 34.9 & 30.5 & 42.8 & 86.8 & - & 88.2 & 66.0 & 34.1 & 86.6 & - & 51.3 & - & 29.4 & 50.5 & 53.3\\
ProDA\cite{zhang2021prototypical} & 93.3 & 61.6 & 85.3 & 19.6 & 5.1 & 37.8 & 36.6 & 42.8 & 84.9 & - & 90.4 & 69.7 & 41.8 & 85.6 & - & 38.4 & - & 32.6 & 53.9 & 55.0\\
DAFormer\cite{hoyer2021daformer} & 84.5 & 40.7 & 88.4 & 41.5 & 6.5 & 50.0 & 55.0 & 54.6 & 86.0 & - & 89.8 & 73.2 & 48.2 & 87.2 & - & 53.2 & - & 53.9 & 61.7 & 60.9\\
DAFormer + \textbf{CLUDA} & 87.4 & 44.8 & 86.5 & 47.9 & \textbf{8.7} & 49.8 & 44.5 & 52.7 & 85.6 & - & 89.2 & 74.4 & 50.2 & 86.9 & - & 65.3 & - & 56.85 & 57.1 & \textbf{61.7} \\
HRDA \cite{hoyer2022hrda} & 85.2 & \textbf{47.7} & 88.8 & \textbf{49.5} & 4.8 & 57.2 & 65.7 & \textbf{60.9} & 85.3 & - & 92.9 & 79.4 & 52.8 & 89.0 & - & 64.7 & - & 63.9 & 64.9 & 65.8 \\
HRDA + \textbf{CLUDA} & \textbf{87.7} & 46.9 & \textbf{90.2} & 49 & 7.9 & \textbf{59.5} & \textbf{66.9} & 58.5 & \textbf{88.3} & - & \textbf{94.6} & \textbf{80.1} & \textbf{57.1} & \textbf{89.8} & - & \textbf{68.2} & - & \textbf{65.5} & \textbf{65.8} & \textbf{66.8} \\
    \bottomrule    
\end{tabular}
\vspace{1em}
\caption{Comparison with state-of-the-art methods for UDA. The results for HRDA + CLUDA are averaged over 3 random seeds with standard deviation of 0.32 and 0.44 for GTA $\rightarrow$ Cityscapes and Synthia $\rightarrow$ Cityscapes respectively.}
\label{tab:uda_comp}
\end{table}
\begin{table}[h]
\setlength\tabcolsep{1.5pt}
        \centering
	\subfloat[\label{tab:compare}]{
            \scriptsize
	    \begin{tabular}{@{}ccc@{}}
            \toprule
            Backbone & DAFormer\cite{hoyer2021daformer} &  DAFormer\cite{hoyer2021daformer} + CLUDA \\ \midrule
            Swin-L\cite{liu2021swin} & 67.4  & \textbf{70.11}   \\
            SegFormer-B5\cite{xie2021segformer} & 68.3  & \textbf{69.9} \\
            ResNet-101\cite{he2016deep} & 56  & \textbf{59.2} \\ \bottomrule
            \end{tabular}
       }
       \qquad
       \subfloat[\label{multi_res_comp}]{
            \scriptsize
            \begin{tabular}{@{}cc@{}}
                \toprule
                Methods & mIoU \\ \midrule
                HRDA\cite{hoyer2022hrda} & 73.8 \\
                HRDA\cite{hoyer2022hrda}+CLUDA (LR only)                & 72.2 \\
                HRDA\cite{hoyer2022hrda}+CLUDA (LR + HR)                & 72.9 \\
                HRDA\cite{hoyer2022hrda}+CLUDA \\(Weighted(LR + HR)) & \textbf{74.4} 
                \\ \bottomrule
            \end{tabular}
        }
\caption{Table (a) shows Comparison of DAFormer\cite{hoyer2021daformer} + CLUDA with DAFormer\cite{hoyer2021daformer} on GTA$\rightarrow$Cityscapes using different backbone network architectures. Table (b) shows comparison of different ways of combining high resolution(HR) and low resolution(LR) features obtained from HR crops and LR crops respectively. In (LR only), the features from LR crops are considered only. In (LR + HR), both LR features and HR features are taken into account and corresponding contrastive losses are added. In (weighted(LR + HR)), we combine the respective contrastive losses by weighing them using a learned feature weight map. The experiments are conducted on GTA $\rightarrow$ Cityscapes.}
\end{table}
\textbf{Training}: We follow the same training regime as followed in DAFormer\cite{hoyer2021daformer} and HRDA\cite{hoyer2022hrda} for DAFormer\cite{hoyer2021daformer} + CLUDA and HRDA\cite{hoyer2022hrda} + CLUDA respectively. However we take $C = 512$ as the dimension of the fused feature map (refer appendix section \ref{tab:compare_embed}). We train our model using AdamW \cite{loshchilov2017decoupled} with learning rate of $6 \times 10^{-5}$ for encoder and a learning rate of $6 \times 10^{-4}$ for decoder and keep betas as $(0.9, 0.999)$, weight decay of 0.01, and a batch size of 2. For self-training we keep the value of EMA weight update parameter $\alpha=0.999$, for learning-rate optimization we follow polynomial-learning rate reduction. We increase the learning-rate for the first 1500 iterations with a warmup rate of $10^{-6}$ and follow the polynomial-reduction of the learning-rate\cite{hoyer2021daformer} after that. In HRDA\cite{hoyer2022hrda} + CLUDA, the learned feature weight obtained from the feature weight prediction head has the dimension $1 \times h \times w$ where $h \times w$ is the spatial dimension of the feature map. We resize the feature map $\mathcal{F}_{fused}$ to $\mathcal{A} = 65 \times 65$. Following \cite{hoyer2021daformer}, we use feature distance loss coefficient $\lambda = 0.005$.
\subsection{Comparison with existing UDA methods}
We begin by comparing the proposed approach with existing UDA methods. We show that CLUDA improves the existing method by a margin of +0.6 mIoU (standard deviation: 0.32) in GTA $\rightarrow$ Cityscapes and +1.0 mIoU (standard deviation: 0.44) in SYNTHIA $\rightarrow$ Cityscapes in Table \ref{tab:uda_comp}.  Class-wise improvements can be seen in 12 of the 19 classes in GTA $\rightarrow$ Cityscapes, Where major improvements can be seen in difficult classes like Sidewalk, Fence, etc., and in 13 of the 16 classes in SYNTHIA $\rightarrow$ Cityscapes, as also supported quantitatively in Table \ref{tab:uda_comp}.
\begin{table}[h!]
    \setlength\tabcolsep{1.5pt}
    \centering
    \subfloat[\label{ablation}]{
    \scriptsize
    \begin{tabular}{@{}cccccc@{}}
        \toprule
        UDA Method & unweighted CL & weighted CL & mIoU \\ \midrule
        DAFormer\cite{hoyer2021daformer}  &               &             & 33.4 \\
        DAFormer\cite{hoyer2021daformer} + CLUDA  & \checkmark   &   & 58.6 \\
        DAFormer\cite{hoyer2021daformer} + CLUDA  &               & \checkmark  & \textbf{69.9} \\ \bottomrule
        \end{tabular}
    }
    \qquad
    \subfloat[\label{stuff_thing_mask_table}]{
    \scriptsize
    \begin{tabular}{@{}ccc@{}}
            \toprule
            CL type & mIoU \\ \midrule
            S-S + T-T + S-T   & 67.4 \\
            S-S       & 68.9 \\
            S-S + T-T   & \textbf{69.9}   \\\bottomrule             
    \end{tabular}
    }    
\caption{Table (a) shows the ablation of weighted and unweighted contrastive loss applied over domain mixed features in DAFormer\cite{hoyer2021daformer} + CLUDA. The contrastive losses are applied over fused feature map obtained from source and domain mixed images separately. However features obtained from domain mixed images are weighted. All experiments are conducted with feature embedding dimension $C=512$. Table (b) shows the ablation of different strategies used while computing contrastive loss over the fused feature maps. We adopt three different strategies and compare the performance of our method. In S-S (stuff-stuff), the contrastive losses are only computed between stuff class features. In S-S + T-T (stuff-stuff + thing-thing) contrastive losses are computed separately over stuff class features and thing class features. In S-S + T-T + S-T (Stuff-Stuff + Thing-Thing + Stuff-Thing), we compute contrastive loss taking into account every pair of feautes irrespective of stuff or thing class. All experiments are conducted on GTA$\rightarrow$Cityscapes. }
\end{table}
\subsection{Comparison of network architectures with CLUDA}
We perform ablation studies with different backbone network architectures in Table \ref{tab:compare}. Our method shows consistent improvement over DAFormer\cite{hoyer2021daformer} across backbone architectures. The best results are obtained with Swin-L\cite{liu2021swin} backbone.           
\subsection{Influence of Confidence Weights on Contrastive loss} \label{sec:conf_weight_sec}
Table \ref{ablation} shows why weighting the contrastive loss computed on the domain-mixed-features is important. Backpropagation signals from the contrastive loss in the early stage of training are very high, when the model predictions are highly unreliable, and are detrimental to the overall performance. Hence, we weight the loss on the teacher-network prediction confidence (equation \ref{eq:conf_computation}). This regularizes the model learning by applying low weights when the prediction confidence is low. The weights increase as the teacher-network becomes more confident. Further, Fig. \ref{gph:pixel_rate} shows how the teacher network makes more confident predictions as the training progresses. The performance drops in the absence of confidence weights, as can be seen in Fig. \ref{gph:weighting_perf_comp}.     
\subsection{Influence of separation of features into stuff-thing classes}\label{Influence of stuff-thing mask}
Transformer-based feature-extractors suffer from shape-bias \cite{bhojanapalli2021understanding,hoyer2021daformer}. We address this issue by applying the contrastive loss separately on stuff class and thing class features in the fused feature map. In other words, contrastive loss is computed taking into account pairwise thing class features and pairwise stuff class features separately, while ignoring stuff-thing feature pairs. In Table \ref{stuff_thing_mask_table}, we compare different strategies used while applying contrastive losses on the features. We acheive the best performance when we treat stuff class features and thing class separately as discussed above. Adopting this strategy reduces the probability of pixels belonging to stuff classes getting misclassified as thing class and vice-a-versa. For eg., The sign boards (part of Building class - stuff class) are often misclassified as traffic signal (thing class). Supporting figure is provided in the appendix (Fig. \ref{fig:yellow_problem}).
\subsection{Qualitative Analysis}
\begin{figure*}
\centering
\includegraphics[width=0.9\linewidth]{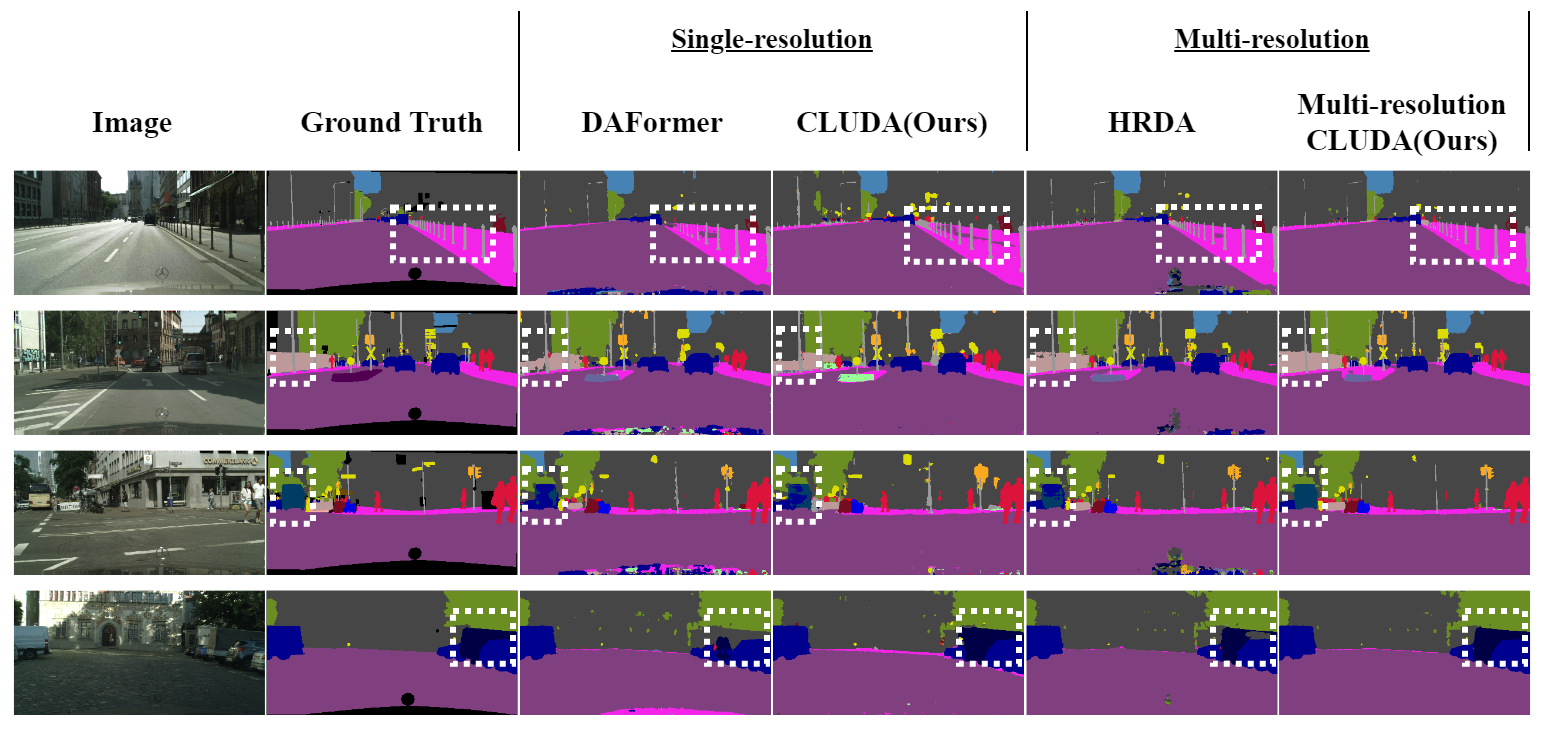}
\caption{ Qualitative analysis of CLUDA on GTA $\rightarrow$ Cityscapes with existing SOTA methods. Note that CLUDA uses only low resolution images and hence must be compared with DAFormer for fair comparison. Multi-resolution CLUDA uses both high and low resolution crops. Fig. shows the performance improvement in classes like Sidewalk, Fence, Bus, etc highlighted using dotted boxes. For more qualitative analysis on GTA $\rightarrow$ Cityscapes and SYNTHIA $\rightarrow$ Cityscapes please refer to appendix (Fig. \ref{fig:more_comparisons}}
\label{img_comp}
\end{figure*}

We provide visual illustrations of how CLUDA improves performance of state-of-the-art UDA methods in semantic segmentation. Fig. \ref{img_comp} shows imporvements achieved by DAFormer\cite{hoyer2021daformer}+CLUDA in single resolution and by HRDA\cite{hoyer2022hrda}+CLUDA in multi-resolution. One can clearly notice the significant improvements in segmentation of difficult classes such as Fence, Sidewalks, etc.
\subsection{Influence of weighting contrastive losses computed on low resolution and high resolution features}
It is important to take into account contrastive losses computed on both low resolution(LR) as well as high resolution(HR) features. However, naively adding the two may not be the best possible strategy to combine the said losses, as can be seen in the Table \ref{multi_res_comp}. We weigh the contrastive losses computed on LR and HR features and add them as per equation \ref{multi_res_cl},  which gives the best performance. Although rows 2 and 3 of Table \ref{multi_res_comp} suggest that using contrastive losses over LR and HR features is not beneficial, by simply letting the model learn the weights for combining losses from LR and HR features gives us a boost in performance.
\section{Conclusion}
In this work, we presented CLUDA, a simple and effective strategy to improve UDA in semantic segmentation using contrastive loss. CLUDA can be used along with existing cross-entropy-based losses to train models. We also showed that CLUDA can be used with different feature-extractor architectures with marginal extra memory consumption. For multi-resolution DA, we extended CLUDA by using learned-feature weighting to combine contrastive loss computed on multiple resolutions. Overall CLUDA produced the state-of-the-art model performance of 74.4 mIoU in GTA $\rightarrow$ Cityscapes with a gain of +0.6 and 67.2 mIoU in SYNTHIA $\rightarrow$ Cityscapes with a gain of +1.4.

\bibliography{egbib}

\newpage

\section{Appendix}

\subsection{Overview}
In this material, we provide additional details on CLUDA and additional experiments and their analysis. We also provide details on datasets and more qualitative analysis with existing state-of-the-art (SOTA) methods. We have organized this material in the following way, section \ref{DATASETS} will have details on dataset we have used in this work, in section \ref{multi-res-da-exp}, we provide more details on experiments mentioned in Table 4b in the main manuscript, section \ref{sec:compare} compares CLUDA results with existing SOTA using different feature encoders, code implementation details are included in section \ref{sec:code_impl}, and section \ref{qualitative_analysis} has further qualitative analysis of CLUDA with exisiting SOTA methods.

\subsection{Datasets}\label{DATASETS}
\textbf{GTA}: The GTA \cite{richter2016playing} dataset has 24,966 synthetic images extracted from a photo-realistic open world game called Grand Theft Auto along with semantic segmentation maps. The image resolution is 1914x1052.

\textbf{SYNTHIA}: \cite{ros2016synthia} is a synthetic dataset that consists of 9400 photo-realistic frames with a resolution of 1280 $\times$ 960, rendered from a virtual city that comes with pixel-level annotations for 13 classes.

\textbf{Cityscapes}: \cite{cordts2016Cityscapes} is a large-scale database that focuses on semantic understanding of urban street scenes. The dataset has semantically annotated 2975 training images and 500 validation images with a resolution of 2048x1024.

\begin{table*}[h!]
\scriptsize
\hspace{-0.8cm}\begin{tabular}{@{}c@{\hspace{0.1cm}}c@{\hspace{0.1cm}}c@{\hspace{0.1cm}}c@{\hspace{0.1cm}}c@{\hspace{0.1cm}}c@{\hspace{0.1cm}}c@{\hspace{0.1cm}}c@{\hspace{0.1cm}}c@{\hspace{0.1cm}}c@{\hspace{0.1cm}}c@{\hspace{0.1cm}}c@{\hspace{0.1cm}}c@{\hspace{0.1cm}}c@{\hspace{0.1cm}}c@{\hspace{0.1cm}}c@{\hspace{0.1cm}}c@{\hspace{0.1cm}}c@{\hspace{0.1cm}}c@{\hspace{0.1cm}}c@{\hspace{0.1cm}}c@{}}
\toprule
 & Road & S.Walk & Build. & Wall & Fence & Pole & T. Light & T. Sign & Veget. & Terrain & Sky & Person & Rider & Car & Truck & Bus & Train & M.Bike & Bike & mIoU \\ \midrule
HRDA \cite{hoyer2022hrda} & 96.4 & 74.4 & 91 & \textbf{61.6} & 51.5 & 57.1 & 63.9 & 69.3 & 91.3 & 48.4 & 94.2 & 79.0 & 52.9 & 93.9 & \textbf{84.1} & 85.7 & \textbf{75.9} & 63.9 & 67.5 & 73.8 \\
Entropy cropping & \textbf{97.3} & \textbf{79.2} & \textbf{91.5} & 61.4 & 53.4 & \textbf{60.6} & 59.2 & \textbf{73.6} & \textbf{91.7} & 51.1 & \textbf{94.7} & \textbf{80.1} & \textbf{54.7} & \textbf{94.5} & 78.2 & 85.1 & 71.7 & 63.9 & \textbf{70.7} & 74.39 \\
Random cropping & 97.1 & 78 & 91 & 60.3 & \textbf{55.3} & 56.3 & \textbf{64.3} & 71.5 & 91.2 & \textbf{51.1} & \textbf{94.7} & 78.4 & 52.9 & \textbf{94.5} & 82.8 & \textbf{86.5} & 73 & \textbf{64.2} & 69.7 & \textbf{74.41} \\ \bottomrule
\end{tabular}
\caption{Comparison of Multi-resolution CLUDA using entropy cropping and random cropping. Here Random cropping is the same experiment mentioned in 10th row in Table 1 of main manuscript(HRDA\cite{hoyer2022hrda} + \textbf{CLUDA(Ours)}). Entropy cropping is HRDA\cite{hoyer2022hrda} + \textbf{CLUDA(Ours)} with entropy cropping instead of random cropping on mixed images as mentioned in section \ref{subsec:entropy_crop}.}
\label{tab:cropping_comp}
\end{table*}

\subsection{Mutli-Resolution DA experiments}\label{multi-res-da-exp}
From Table 4b in the main manuscript, One can see that contrastive loss on just the low-resolution image (LR only in Table 4b) is ineffective. This may be due to the reason that the cross-entropy (CE) is forcing the model to learn a different representation compared to the contrastive loss. To further examine this, we conduct an experiment where we simply add the contrastive loss computed on low-resolution and high-resolution image crops (LR + HR in Table 4b) as shown in equation \ref{multi_res_cl}. As expected, this experiment yielded better results compared to LR only but is not as good as the results obtained when we use learned-weighted addition (Row 3 in Table 4b in main manuscript).

\begin{equation}
\label{multi_res_cl}
\begin{split}
& \mathcal{L}_{CL}^{MR}(x^{l}_m, x^{h}_m, \mathcal{A}^{s}, \mathcal{A}^{m}) = 
\frac{1}{\mathcal{A}} \biggl[ \sum\limits_{s \in \mathcal{S}_l} \delta_l \mathcal{L}^{I}_{NCE}(\vec{v}^{l}_s, \mathcal{P}^{l}_{s}, \mathcal{N}^{l}_s) + \sum_{s\in S_h} (1 - \delta_l)  \mathcal{L}_{NCE}^I (\vec{v}^{h}_s, \mathcal{P}^{h}_{s}, \mathcal{N}^{h}_{s} ) \\ 
& \qquad + \Gamma \sum_{t \in \mathcal{T}_l} \delta_{l} \mathcal{L}^{I}_{NCE}(\vec{v}^{l}_{t}, \mathcal{P}^{l}_{s,t}, \mathcal{N}^{l}_{s,t}) + \Gamma \sum_{t\in \mathcal{T}_h} (1 - \delta_l)  \mathcal{L}_{NCE}^I (\vec{v}^{h}_{t}, \mathcal{P}^{h}_{s,t}, \mathcal{N}^{h}_{s,t} ) \biggr]. 
\end{split}
\end{equation}

\textbf{Note} the omission of the $\delta$ hyper-parameter in equation \ref{multi_res_cl}. This is in line with the findings in \cite{hoyer2022hrda}, where the authors have used weighted addition of CE loss computed on different resolutions. To improve the results further, taking inspiration from \cite{hoyer2022hrda}, we proposed learned-weighted addition of contrastive loss computed on low-resolution and high-resolution image crops. For this, we use the same simple, context fusion decoder that is used to predict segmentation label \cite{hoyer2021daformer, hoyer2022hrda} but alter the final layer to predict a single channel weight. The predicted weights are then used to compute the weights of the pixel contrastive loss  in equation 5 in main manuscript. We use the confidence based re-weighting, $ \Gamma_{conf} $, of contrastive losses computed on domain mixed images.


\subsection{Multi-Resolution DA with entropy based cropping}\label{subsec:entropy_crop}
\begin{figure}
    \centering
    \includegraphics[width=0.9\linewidth]{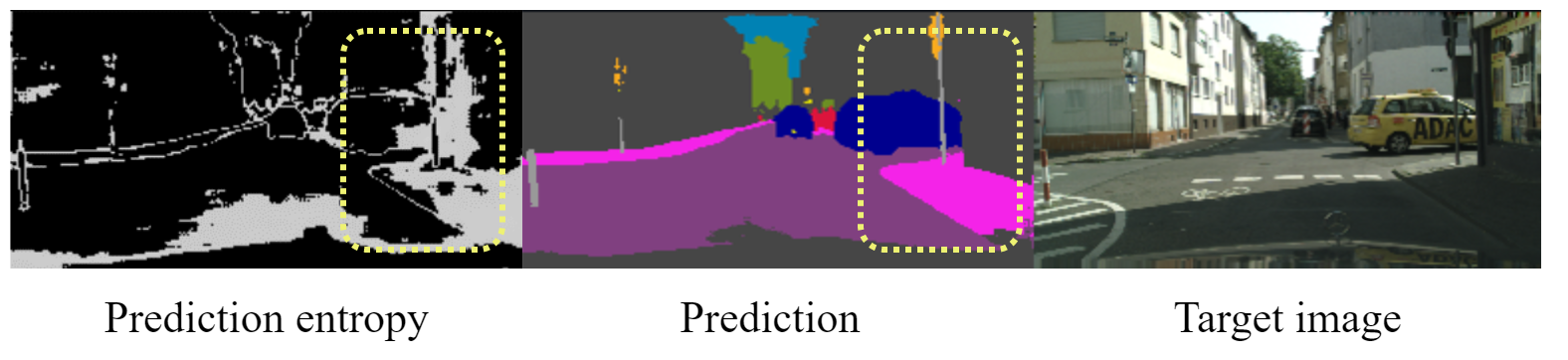}
    \caption{This figure shows how entropy varies spatially over the predicted label.}
    \label{fig:entropy_pred}
\end{figure}

Semantic segmentation labels are often predicted with uneven confidence spatially, i.e some regions are predicted with low confidence while other regions of the same image are predicted with high confidence. See Fig. \ref{fig:entropy_pred}. This leads to poor prediction performance in those regions. We posit that this can be resolved to some extent if the model learns from high resolution crop of these regions. We propose entropy based cropping, i.e we create multiple regional crops from the entropy map of the prediction. For this, we follow the cropping approach mentioned in \cite{hoyer2022hrda}, we use a sliding window of size $(H/\mu \times W/\mu)$ and stride of $(H/4 \times W/4)$, where $(H \times W)$ is the size of the entropy map, to create multiple overlapping regional crops. We compute the mean entropy of each regional crop and select the bounding box that encloses the region with maximum mean entropy. This bounding box is then used for cropping the high-resolution target image and its corresponding pseudo label. \textbf{Note} that this cropping process is applied only while training with domain mixed images, for training with source images we follow random cropping. We aim to improve target prediction by making the model learn from high resolution crops of regions where it predicts poorly. Rest of the training paradigm remains same as \cite{hoyer2022hrda}. Entropy based cropping with our CLUDA gave encouraging results as can be seen in Table \ref{tab:cropping_comp}.

\subsection{Comparison of CLUDA with SOTA using different styles of feature encoders}\label{sec:compare}
In Table 3b of main manuscript, we presented the results of CLUDA on different types of feature encoders. In this section we further compare CLUDA results with previous SOTA using different types of feature encoders. See Table \ref{tab:compare}. One can clearly see that CLUDA outperforms existing SOTA irrespective of the style of feature encoder used.

\subsection{Influence of Embedding Size on Contrastive Loss} \label{sec:embed_size_sec}
Table \ref{tab:compare_embed} shows the improvement in performance because of a larger embedding size. \cite{hoyer2021daformer, hoyer2022hrda} uses an embedding size $C=256$. Table \ref{ablation} shows that merely increasing the embedding size will not give better results in CE-based experiments, as in  \cite{hoyer2021daformer}. But along with the contrastive loss, the increase in $C$ shows improvement in the performance. The performance improvement plateaus at $C=768$. We chose $C=512$ as we find this choice to be a good trade-off between performance, training time, and GPU constraints.

\begin{table}[]
    \scriptsize
    \centering
    \begin{tabular}{@{}ccc@{}}
            \toprule
            Backbone & C    & mIoU \\ \midrule
            Swin-L\cite{liu2021swin}   & 256  & 69   \\
            Swin-L\cite{liu2021swin}   & 512  & 69.2 \\
            Swin-L\cite{liu2021swin}   & 768  & \textbf{69.3} \\
            Swin-L\cite{liu2021swin}   & 1024 & 68.9 \\ \bottomrule
            \end{tabular}
    \vspace{1em}
    \caption{Comparison of CLUDA with existing SOTA using different feature encoders.}
    \label{tab:compare_embed}
\end{table}

\subsection{Code implementation}\label{sec:code_impl}
Code implementation will be available at \url{https://github.com/user0407/CLUDA}.

\subsection{Qualitative Analysis}\label{qualitative_analysis}
\begin{figure*}
    \centering
    \includegraphics[width=1\linewidth]{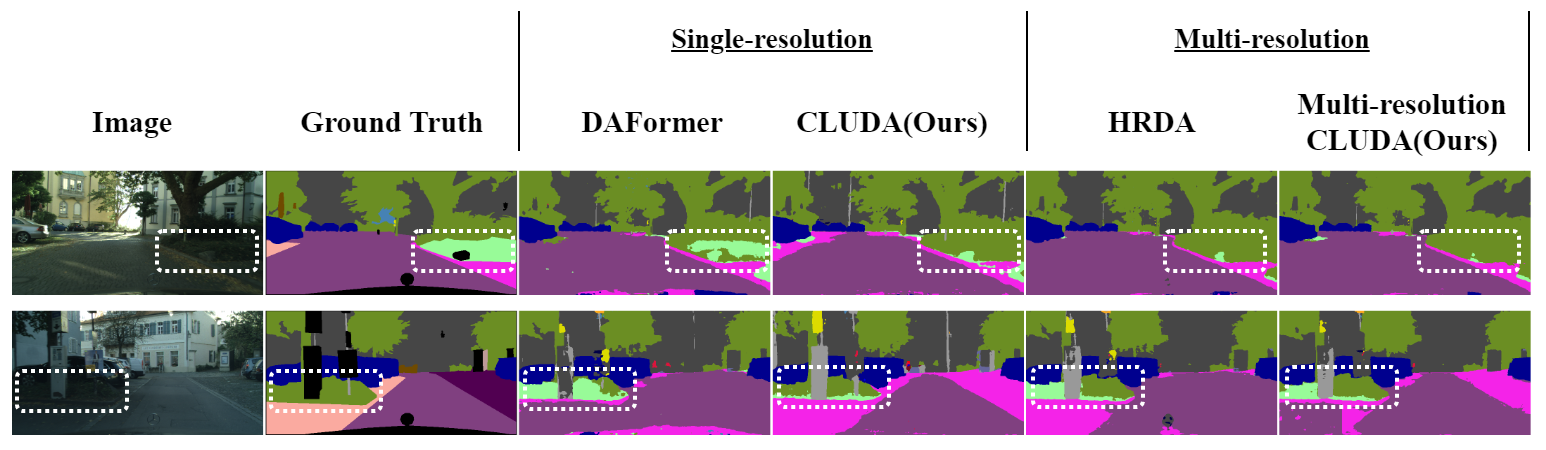}
    \caption{Failure cases due to annotation differences. In this image one can see that the same type of bush, i.e same texture and same height, are annotated as Vegetation (top row) and Terrain (bottom row).}
    \label{fig:green_problem}
    \includegraphics[width=1\linewidth]{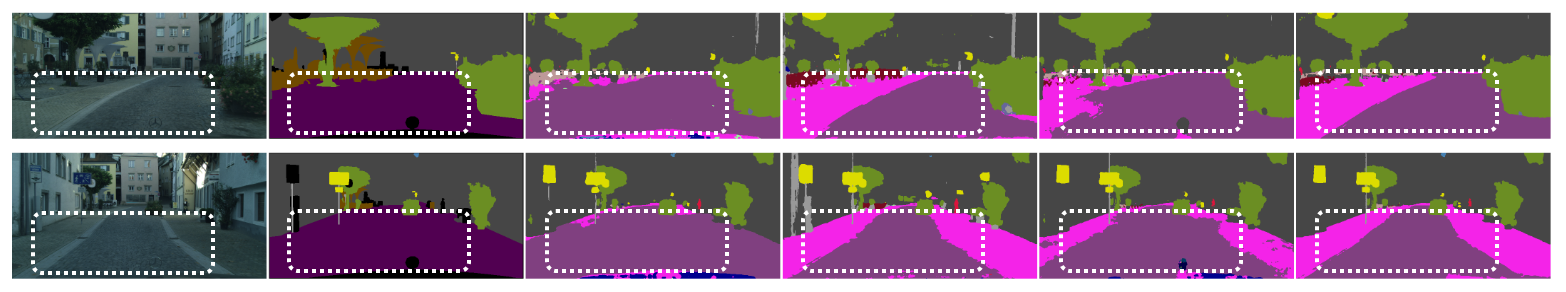}
    \caption{Several pavements that look like sidewalk are annotated as Roads, especially the cobblestone roads. Although CLUDA improves the Road and Sidewalk predictions, these annotation differences are unfavourable for robust predictions.}
    \label{fig:red_problem}
    \includegraphics[width=1\linewidth]{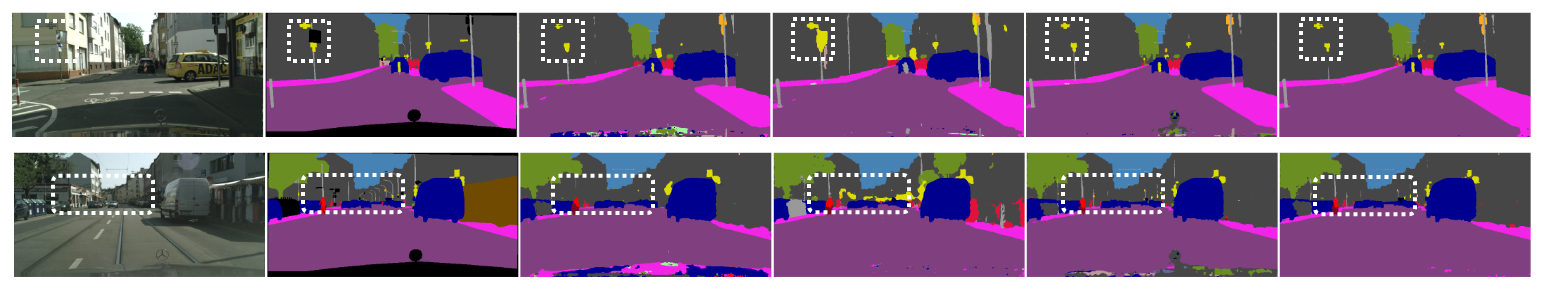}
    \caption{Semantic predictions of Traffic sign are poor compared to existing SOTA, refer to Table 1 in main manuscript. As can be seen from the image, the model is getting confused between Traffic signs and advertisement banners/boards hanging on shops.}
    \label{fig:yellow_problem}
\end{figure*}

In this section we show more comparisons of our CLUDA with existing state of the art methods. In Fig.\ref{fig:green_problem} and Fig.\ref{fig:red_problem}, we show how inconsistencies in annotation policies are affecting the semantic segmentation predictions. In Fig. \ref{fig:yellow_problem}, we mention the failure case of Traffic sign getting predicted poorly. We posit that this may be due to model getting confused between Traffic signs and advertisement banners/boards hanging on shops.

\begin{figure*}
    \centering
    \includegraphics[width=1\linewidth]{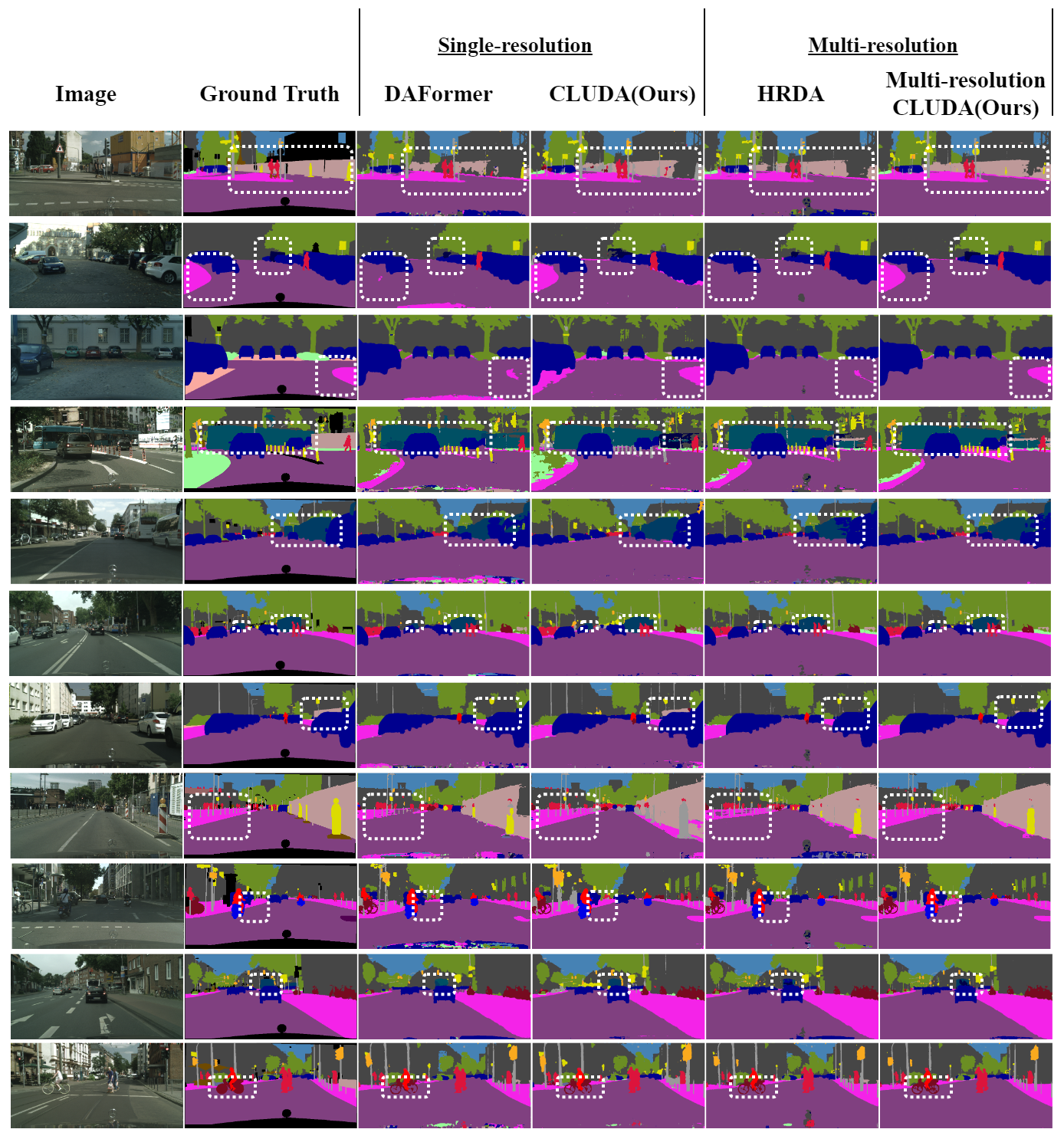}
    \caption{Comparison of CLUDA with existing state of the art methods on GTA $\rightarrow$  Cityscapes.}
    \label{fig:more_comparisons}
\end{figure*}

\end{document}